\documentclass[10pt,twocolumn,letterpaper]{article}

\usepackage{iccv}      
\usepackage{color}
\usepackage{tabularx}
\usepackage{booktabs}
\usepackage{algorithm}
\usepackage{algorithmicx}
\usepackage{algpseudocode}  
              
\usepackage{times}
\usepackage{epsfig}
\usepackage{graphicx}
\usepackage{amsmath}
\usepackage{amssymb}
\usepackage{enumitem}
\usepackage{xspace}
\usepackage{balance}
\usepackage{multirow}
\usepackage{mathtools}
\usepackage{rotating}
\usepackage{url}
\usepackage{tabulary}
\usepackage[table]{xcolor}
\usepackage{array}
\usepackage{subcaption}
\usepackage[export]{adjustbox}
\usepackage{ctable}
\newcolumntype{Y}{>{\centering\arraybackslash}X}
               
\usepackage[pagebackref=true,breaklinks=true,letterpaper=true,colorlinks,bookmarks=false]{hyperref}

\newcommand{\fig}{{Figure}\@\xspace}
\newcommand{\tab}{{Table}\@\xspace}

\def\Vec#1{{\boldsymbol{#1}}}
\def\Mat#1{{\boldsymbol{#1}}}

\iccvfinalcopy % *** Uncomment this line for the final submission

\def\etc{\emph{etc.}}

\def\iccvPaperID{1163} % *** Enter the ICCV Paper ID here

\graphicspath{{./figs/}}
 
\begin{document}

\title
	{
	Dual-Glance Model for Deciphering Social Relationships 
	}

\author{Junnan Li\\
	Graduate School for Integrative Sciences and Engineering\\
	National University of Singapore\\
	Singapore\\
	{\tt\small lijunnan@u.nus.edu}
	\and
	Yongkang Wong\\
	Interactive \& Digital Media Institute\\
	National University of Singapore\\
	Singapore\\
	{\tt\small yongkang.wong@nus.edu.sg}
	\and
	Qi Zhao\\
	Department of Computer Science and Engineering\\
	University of Minnesota\\
	Minneapolis, USA\\
	{\tt\small qzhao@cs.umn.edu}
	\and
	Mohan S. Kankanhalli\\
	School of Computing\\
	National University of Singapore\\
	Singapore\\
	{\tt\small mohan@comp.nus.edu.sg}
}   
            
\maketitle

\label{sec:abstract}
\begin{abstract}

Since the beginning of early civilizations,
social relationships derived from each individual fundamentally form the basis of social structure in our daily life.
In the computer vision literature, 
much progress has been made in scene understanding,
such as object detection and scene parsing.
Recent research focuses on the relationship between objects based on its functionality and geometrical relations.
In this work, 
we aim to study the problem of social relationship recognition,
in still images.
We have proposed a dual-glance model for social relationship recognition,
where the first glance fixates at the individual pair of interest and the second glance deploys attention mechanism to explore contextual cues.
We have also collected a new large scale People in Social Context (PISC) dataset, 
which comprises of 22,670 images and 76,568 annotated samples from 9 types of social relationship.
We provide benchmark results on the PISC dataset, 
and qualitatively demonstrate the efficacy of the proposed model.

\end{abstract}

\section{Introduction}
\label{sec:introduction}

%Since the beginning of early civilizations,
Social relationships derived from each individual fundamentally form the basis of social structure in our daily life.
%Today,
%being socially connected is a basic need for human beings,
%where we interact with the world through the network of people connected to us.
Naturally, 
we perceive and interpret a scene with an understanding of the social relationships of the people in the scene.
Sociology research shows that such social understanding of people permits inference about their characteristics and their possible behaviors~\cite{Smith_SC_1990}.
\begin{figure}[!t]
  \centering
  \begin{minipage}{1.0\columnwidth}
  	\begin{minipage}{1.0\columnwidth}
  		\begin{minipage}{0.495\columnwidth} \centerline{\includegraphics[width=\linewidth]{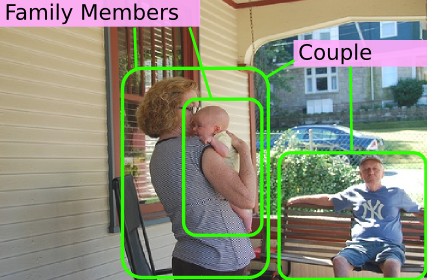}} 			\end{minipage}
  		\begin{minipage}{0.495\columnwidth} \centerline{\includegraphics[width=\linewidth]{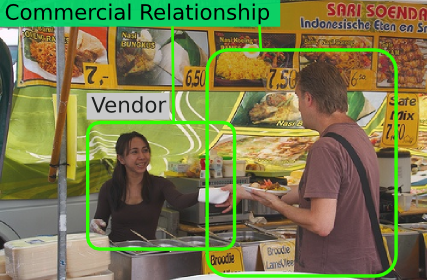}} 	\end{minipage}
  	\end{minipage}
  	\begin{minipage}{1.0\columnwidth}
  		\begin{minipage}{0.495\columnwidth} \centerline{\includegraphics[width=\linewidth]{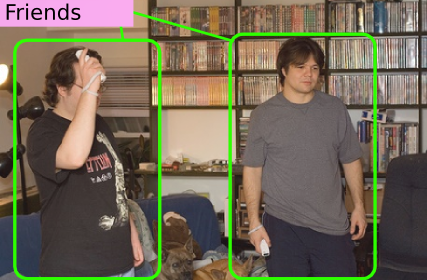}} 			\end{minipage}
  		\begin{minipage}{0.495\columnwidth} \centerline{\includegraphics[width=\linewidth]{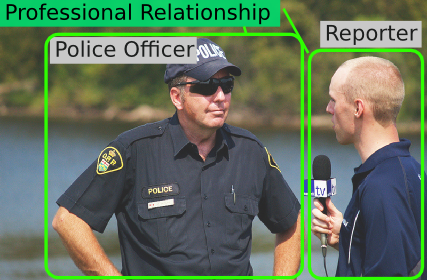}}	\end{minipage}
  	\end{minipage}
  	\begin{minipage}{1.0\columnwidth}
  		\begin{minipage}{0.495\columnwidth} \centerline{\includegraphics[width=\linewidth]{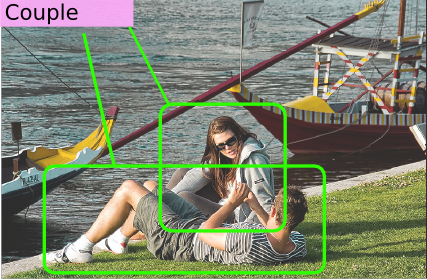}} 			\end{minipage}
  		\begin{minipage}{0.495\columnwidth} \centerline{\includegraphics[width=\linewidth]{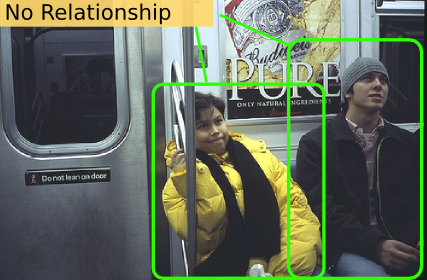}} 					\end{minipage}
  	\end{minipage}
  \end{minipage}
%   \begin{minipage}{1.0\columnwidth}
%   	\begin{minipage}{1.0\columnwidth}
%   		\begin{minipage}{0.327\columnwidth} \centerline{\includegraphics[width=\linewidth]{PISC_sample_family}} 			\end{minipage}
%   		\begin{minipage}{0.327\columnwidth} \centerline{\includegraphics[width=\linewidth]{PISC_sample_friends}} 			\end{minipage}
%   		\begin{minipage}{0.327\columnwidth} \centerline{\includegraphics[width=\linewidth]{PISC_sample_couple}} 			\end{minipage}
%   	\end{minipage}
%   	\begin{minipage}{1.0\columnwidth}
%   		\begin{minipage}{0.327\columnwidth} \centerline{\includegraphics[width=\linewidth]{PISC_sample_commercial}} 	\end{minipage}
%   		\begin{minipage}{0.327\columnwidth} \centerline{\includegraphics[width=\linewidth]{PISC_sample_professional}}	\end{minipage}
%   		\begin{minipage}{0.327\columnwidth} \centerline{\includegraphics[width=\linewidth]{PISC_sample_no}} 					\end{minipage}
%   	\end{minipage}
%   \end{minipage}
  \vspace{-1.5ex}
  \caption
    {
    \small
		Example images from the new People in Social Context (PISC) dataset.
    } 
  \label{fig:image_example}
  \vspace{-2ex}
\end{figure}

In computer vision, 
social information has been exploited to improve several analytics tasks, 
including human trajectory prediction~\cite{Alahi_2016_CVPR,Robicquet_ECCV_2016}, 
multi-target tracking~\cite{Choi_ECCV_2012,Qin_CVPR_2012},
and group activity recognition~\cite{Direkoglu_ECCV_2012,Lan_CVPR_2012,Lan_PAMI_2012}.
In image understanding task, 
visual concepts recognition is gaining more attention,
which include visual attribute~\cite{Huang_2016_CVPR} and visual relationship~\cite{Lu_ECCV_2016}.
On the other hand,
social attribute and social relationship~\cite{Wang_ECCV_2010} are equally important concepts for scene understanding, 
but have received less attention in the research community.
%Therefore, 
%we posit that there exists a gap between machine and human in the perception of people in a scene.
%We refer to this as {\it social gap}.
In this work,
we aim to address the problem of social relationship recognition.
Understanding such relationship can enable a well designed algorithm to generate better descriptions for a scene. 
For instance, 
the first image in \fig~\ref{fig:image_example} can be described as `Grandma is holding her grandchild', rather than `A person is holding a baby'.
%In addition, robots can better interact with humans if they can interpret the social meaning of a scene.

\begin{figure}[!t]
	\centering
	\begin{minipage}{1.0\columnwidth}
		\centerline{\includegraphics[width=1.0\linewidth]{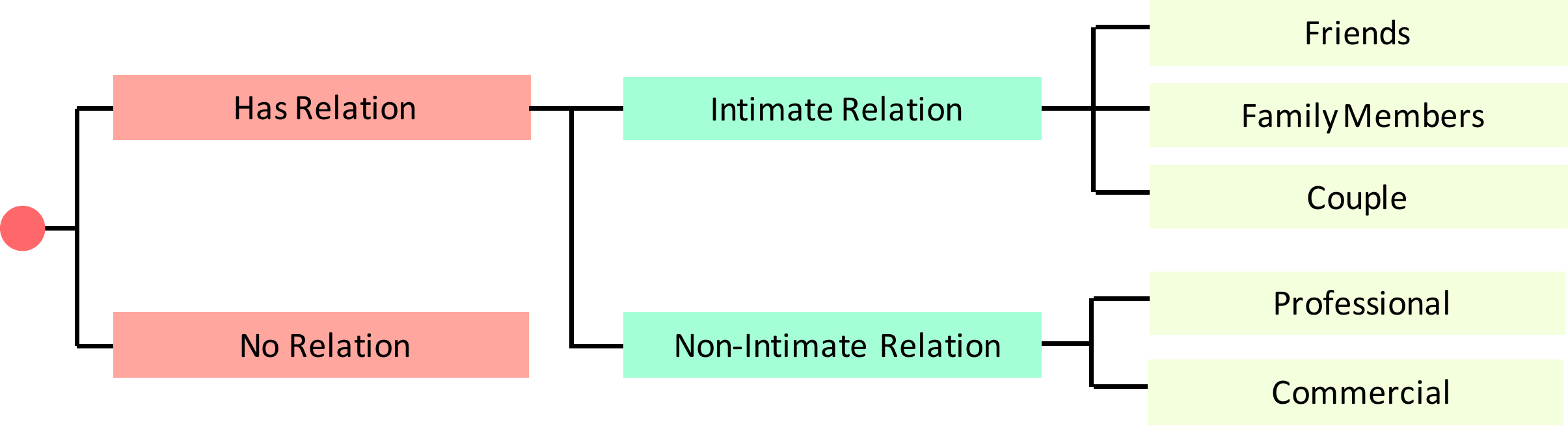}}
	\end{minipage}
  \vspace{-1ex}
  \caption
    {
    \small     
    Defined hierarchical social relationship categories.
    }
 	\vspace{-2ex}
  \label{fig:social_relationship_tree}
\end{figure}

With reference to the {\it relational models} theory~\cite{Fiske_PR_1992},
we define a hierarchical social relationship categories which embed the coarse-to-fine characteristic of common social relationships
(as illustrated in \fig~\ref{fig:social_relationship_tree}).
Our definition follows a prototype-based approach, 
where we are interested in finding exemplars that most parsimoniously describe the most common situations,
rather than an ambiguous definition that could cover all possible cases.
The presented recognition problem differs from the visual relationship detection problem in \cite{Lu_ECCV_2016}.
We argue that inferring social relationship requires a higher level of understanding about the scene.
This is because humans make such inferences not only based on the physical appearance (\eg,~color of clothes, gender, age, \etc),
but also from subtler cues (\eg,~expression, action, proximity, and context)~\cite{Alletto_CVPR_2014,Ramanathan_CVPR_2013,Zhang_ICCV_2015}.

Recognizing social relationships from still images is challenging due to the wide variations in scale, scene, pose, and appearance.
In this work, 
we propose a dual-glance model, 
which exploits information from a target individual pair as well as the surrounding contextual cues.
The key contributions can be summarized as:
\begin{itemize}
  \item 
  The proposed dual-glance model mimics human visual system to explore useful and complementary visual cue for social relationship analysis.
  The first glance fixates at the individual pair of interest, 
	and performs coarse prediction based on its appearance and geometrical information.
	The second glance exploits contextual cues from regions generated from Region Proposal Network (RPN)~\cite{Ren_NIPS_2015} to refine the coarse prediction.
  \item 
  We propose Attentive RCNN, where attention is allocated for each contextual region.
  The attention mechanism is guided by both bottom-up and top-down signals.
  Better performance is achieved by selectively focusing on the relevant regions.
  \item 
  To enable this study, 
	we collected a novel People in Social Context (PISC) dataset\footnote{\url{https://doi.org/10.5281/zenodo.832013}}.
	It consists of 22,670 images and 76,568 manually annotated labels from 9 types of social relationship.
% 	Example images are shown in \fig~\ref{fig:image_example}.
	In addition, 
	PISC also consists of 66 annotated occupation categories.
	To the best of our knowledge, 
	PISC is the first public dataset for social relationship analysis.
\end{itemize}

The remaining of the paper is organized as follows.
Section~\ref{sec:literature} reviews the related work.
Section~\ref{sec:dataset} delineates the details of the new PISC dataset. 
Section~\ref{sec:model} elaborates on the details of the proposed framework, 
and the empirical evaluation is shown in Section~\ref{sec:experiment}.
Section~\ref{sec:conclusion} concludes the paper.

\section{Related Work}
\label{sec:literature}
 
\subsection{Social Relationship}
The study of social relationships lies at the heart of social sciences.
%Social relationships are the cognitive sources for generating social action, 
%for understanding individual's social behavior, and for coordinating social interaction~\cite{Haslam_JESP_1992}.
There are two forms of representations for relational cognition. 
The first approach represents relationship with a set of theorized or empirically derived dimensions~\cite{Conte_JPSP_1981}.
The other form of representation proposes implicit categories for relation cognition~\cite{Haslam_C_1994}.
One of the most extensively accepted categorical theory is the relational models theory~\cite{Fiske_PR_1992}.
It offers a unified account of social relations by proposing four elementary prototypes,
namely \textit{communal sharing}, \textit{equality matching}, \textit{authority ranking}, and \textit{market pricing}.
%The relational models theory is unique among theories of social relationships because it unambiguously represents 
%discrete relationship frames that are universal cultural.
 
In the computer vision literature, 
social information has been widely adopted as supplementary cues to other tasks.
Gallagher~\etal~\cite{Gallagher_CVPR_2009} extract features describing group structure to aid demographic recognition.
For group activity recognition,
social roles and relationship information have been implicitly embedded into the inference model~\cite{Choi_ECCV_2012,Deng_CVPR_2016,Direkoglu_ECCV_2012,Lan_CVPR_2012,Lan_PAMI_2012}.
Alletto~\etal~\cite{Alletto_CVPR_2014} define `social pairwise feature' based on F-formation and use it for group detection in egocentric videos. 
Recently, 
\cite{Alahi_2016_CVPR,Robicquet_ECCV_2016} model social factor for human trajectory prediction.

There have been studies that explicitly focus on recognition of social attributes and social structures. 
%(\ie,~birthday, wedding, award function, physical training).
Wang~\etal~\cite{Wang_ECCV_2010} first study familial social relationship recognition in personal image collections.
Kinship verification~\cite{Hamdi_ICCV_2013,Fang_ICIP_2010,Xia_TMM_2012} and kinship recognition~\cite{Chen_ACMMM_2012,Guo_ICPR_2014} have been extensively studied.
Zhang~\etal~\cite{Zhang_ICCV_2015} study facial traits (\eg,~friendly, dominant, \etc) that are informative of social relationships.
For video based analysis,
Ding and Yilmaz discover social communities formed by actors in movies~\cite{Ding_HCSMA_2014}.
Ramanathan~\etal~\cite{Ramanathan_CVPR_2013} study weakly supervised social role discovery in events. 
%On occupation recognition problem,
%\cite{Song_ICCV_2011} propose a part-based representation of human clothing for predicting occupation of a single person. 
%\cite{Shao_ICCV_2013} apply occupation recognition to multiple people scenario with arbitrary poses.
%In both cases, 
%no more than 20 occupations were considered.
 
Our study partially overlaps with the field of social signal processing~\cite{Vinciarelli_TAC_2012},
which aims to understand social signals and social behaviors using multiple sensors,
such as role recognition, influence ranking, and dominance detection in group meeting~\cite{Hung_ACMM_2007,Rienks_ICMI_2006,Salamin_TMM_2009}.
%as well as recognition of social emotions~\cite{Gunes_IJSE_2010}, \eg~empathy, envy and admiration.
% 
Our work substantially differs from the aforementioned studies. 
Unlike facial attributes based social relationship study~\cite{Chen_ACMMM_2012,Guo_ICPR_2014,Wang_ECCV_2010,Zhang_ICCV_2015},
we study people in complex daily scenes with uncontrolled poses and orientations.
Furthermore, 
we focus on general social relationships, 
rather than kinship in family photos~\cite{Chen_ACMMM_2012,Hamdi_ICCV_2013,Fang_ICIP_2010,Wang_ECCV_2010,Xia_TMM_2012}.
Different from video-based studies~\cite{Ding_HCSMA_2014,Ramanathan_CVPR_2013},
% Compared with~\cite{Ding_ICCV_2011,Ramanathan_CVPR_2013} that studied videos,
we focus on visual information from a single image.

\subsection{Multiple-Instance Learning} 

The proposed Attentive RCNN is inspired by Multiple-Instance Learning (MIL).
MIL is a weakly-supervised learning approach which trains a classifier with bags of instances and bag-level labels.  
% where the learner receives bags of instances with bag-level labels.
%which the learner receives positive and negative bags of instances,
%and the bags is considered positive if it contain at least one positive sample~\cite{Maron_NIPS_1997}. 
% It has been widely applied in computer vision tasks. 
Recently, 
researchers explored MIL with deep feature representations. 
%MIL is incorporated with Convolutional Neural Networks (CNNs) for weakly-supervised object localization~\cite{Song_NIPS_2014} and object segmentation~\cite{Pedro_CVPR_2015}, where only image-level labels are available.
Wu~\etal~\cite{Wu_CVPR_2015} propose a deep MIL framework to exploit correspondences between keywords and image regions for image classification and annotation,
while a similar technique was adopted to detect salient concepts for image captions generation~\cite{Feng_CVPR_2015}.
Inspired by MIL,
Gkioxari~\etal~\cite{Gkioxari_ICCV_2015} propose R*CNN.
Different from previous approaches,
it localizes target region for action recognition by exploiting complementary representative cue from a set of candidate regions in an image.
%  selects the most informative contextual region from a set of candidate secondary regions for action recognition.
% which incorporates Fast R-CNN~\cite{Girshick_ICCV_2015} to extract a set of candidate secondary regions,
% followed by selects the most informative region for action recognition.

Attention model has been recently proposed and applied to image captioning~\cite{XU_ICML_2015,You_CVPR_2016}, image question answering~\cite{Yang_2016_CVPR} 
and fine-grained classification~\cite{Xiao_CVPR_2015}.
We modify R*CNN with attention mechanism to better exploit contextual cues.
We treat the attention weights for the contextual regions as latent variable, which can be inferred with a forward pass of the model.
 
\section{People in Social Context Dataset}
\label{sec:dataset}

The People in Social Context (PISC) dataset is the first of its kind that focuses on social relationships.
It was collected through a pipeline of three stages. 
In the first stage, 
we collected around 40k images containing people from a variety of sources, 
including Visual Genome~\cite{Krishna_CoRR_2016}, MS-COCO~\cite{Lin_ECCV_2014}, YFCC100M~\cite{Thomee_ACMC_2016},
Flickr, Instagram, Twitter and commercial search engines (\ie~Google and Bing).
We used a combination of key words search (\ie~co-worker, people, friends, \etc) and people detector (Faster RCNN~\cite{Ren_NIPS_2015}) to collect the image.
The collected images have high variation in image resolution, people's appearance, and scene type.

\begin{table}[!t]
	\centering
	\caption
		{
		\small	
		Instructions provided to annotators.
		}
	\label{tbl:annotation}
	\vspace{-2ex}
	\resizebox{\columnwidth}{!}{
	\begin{tabular}{l|l|l} 
	  \toprule
	  Relationship & Description                            & Examples		\\
	  \midrule
	  Professional & The people are related based on   & co-worker; coach \& player;  \\
	               & their professions                      & boss \& staff\\
	  Commercial   & One person is paying money to receive  & salesman \& customer; \\
	               & goods/service from the other           & tour guide \& tourist\\
	  \bottomrule
	\end{tabular}
	}
	\vspace{-2ex}
\end{table}	

\begin{figure}[!t]
% 	\captionsetup[subfigure]{labelformat=empty}
%   \centering
% 	\begin{subfigure}[t]{0.28\columnwidth}
% 		\centering{\includegraphics[height=0.83in]{unsure_1}}
% 		\caption{\footnotesize Friends? Family?}
% 		\label{}
% 	\end{subfigure}
% 	\begin{subfigure}[t]{0.36\columnwidth}
% 		\centering{\includegraphics[height=0.83in]{unsure_2}}
% 		\caption{\footnotesize Family? Couple?}
% 		\label{}
% 	\end{subfigure}
% 	\begin{subfigure}[t]{0.32\columnwidth}
% 		\centering{\includegraphics[height=0.83in]{unsure_3}}	
% 		\caption{\footnotesize  Friends? Co-workers?}
% 		\label{}
% 	\end{subfigure}
	\centering
	\begin{minipage}{0.95\columnwidth}
		\centerline{\includegraphics[width=1.0\linewidth]{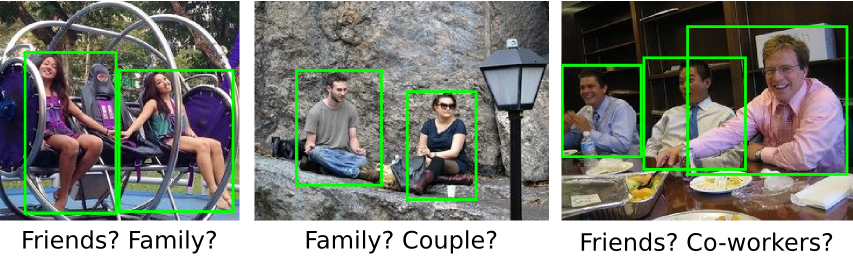}}
	\end{minipage}	
  \vspace{-2ex}
	\caption
		{
		\small  
		Example of social relationship labels that are not agreed among annotators.
		}
  \label{fig:unsure_example}
  	\vspace{-2ex} 
\end{figure}

In the second and third stage, 
we hired workers from CrowdFlower platform to perform labor intensive task of manual annotation.
The second stage focused on the annotation of person bounding box in each image. 
Following \cite{Krishna_CoRR_2016}, 
each bounding box is required to strictly satisfy the coverage and quality requirements. 
To speed up the annotation process, 
we first deployed Faster RCNN to detect people on all images, 
followed by asking the annotators to re-annotate the bounding boxes if the computer-generated bounding boxes were inaccurately localized.
Overall, 
40\% of the computer-generated boxes are kept. 
For images collected from MSCOCO and Visual Genome, 
we directly used the provided groundtruth bounding boxes.

Once the bounding boxes of all images had been annotated, 
we selected images consisting of at least two people who occupy a significant amount of region,
and avoided images that contain crowds of people where individuals cannot be distinguished.
In the final stage,
we requested the annotators to identify the occupation of all individuals in the image,
as well as the social relationships of all potential individual pairs.
To ensure consistency in the occupation categories,
the annotation is based on a list of reference occupation categories.
The annotators could manually add a new occupation category if it was not in the list.

For social relationships, 
we formulate the annotation task as multi-choice questions based on the hierarchical structure in \fig~\ref{fig:social_relationship_tree}. 
We provide instructions (see~\tab~\ref{tbl:annotation}) to help the annotators distinguish between professional and commercial relationship.
Annotators can choose the option `not sure' at any level if they cannot confidently identify the relationship.
Each image was annotated by five workers, and the final decision is determined by majority voting.
If the five workers do not reach an agreement (e.g. 2-2-1), 
the annotation will be treated as invalid 
(see \fig~\ref{fig:unsure_example}).
Overall, 
7,928 unique workers have contributed to the annotation.

\begin{figure}[!t]
  \centerline{\includegraphics[width=1.0\columnwidth]{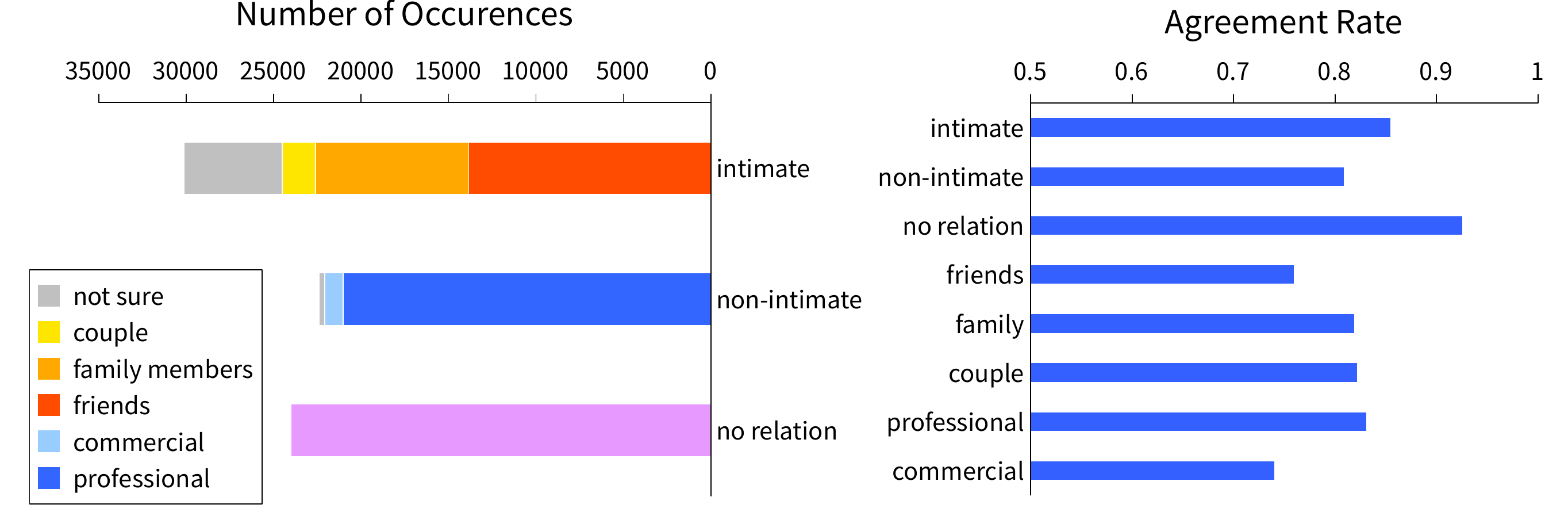}}
  \vspace{-2ex}
	\caption
	  {
	  \small
	  Annotation statistics of the relationship categories.
% 	  Number of occurrences and agreement rate of different categories of relationships.	
	  }  
	\vspace{-2ex} 
  \label{fig:relation_number}
\end{figure}
\begin{figure}[!t]
  \centerline{\hspace*{-2ex}\includegraphics[width=0.83\columnwidth]{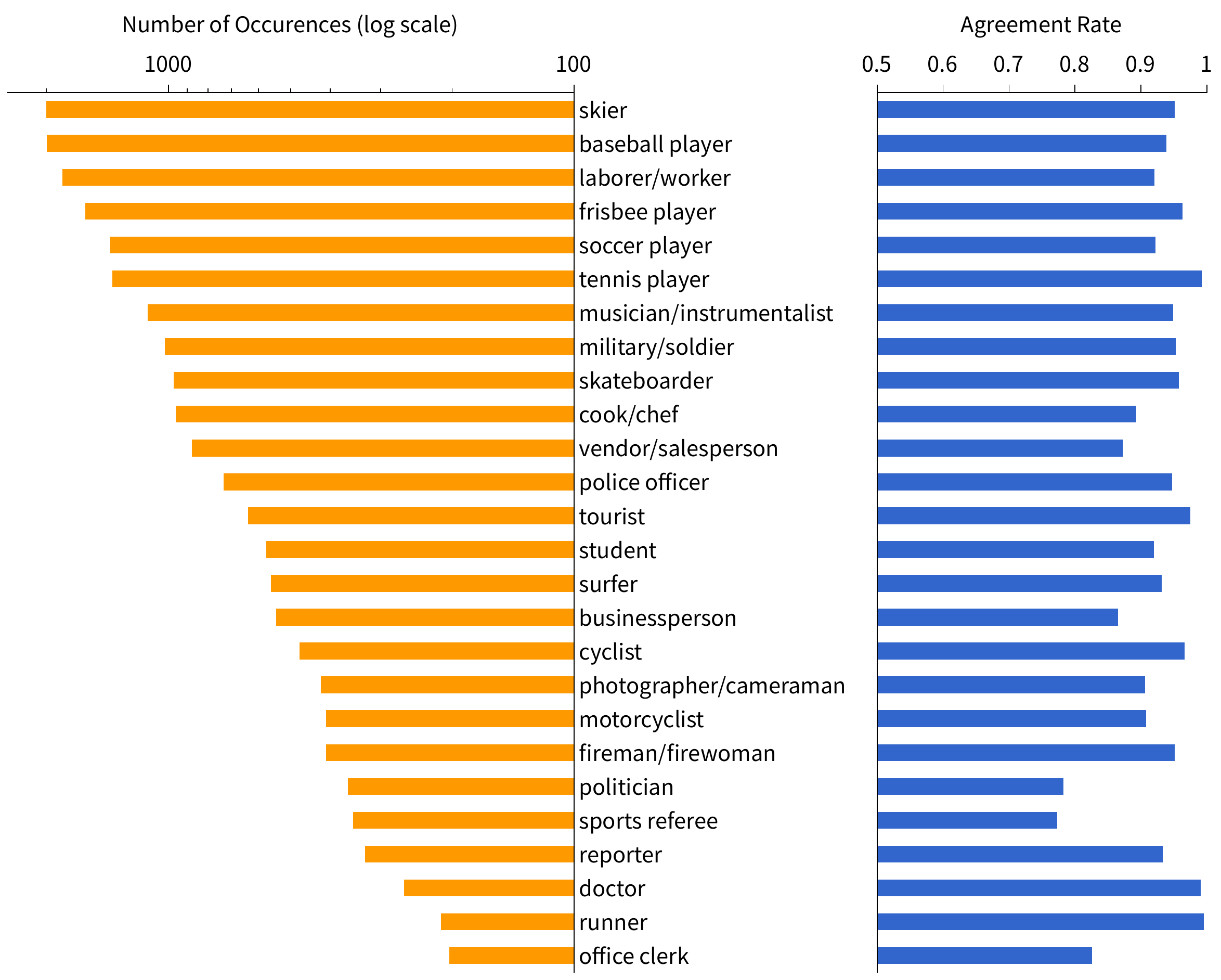}}
  \vspace{-2ex}
  \caption
	  {
	  \small
	  Annotation statistics of the top 26 occupations.
% 		Annotation statistics of the top 26 occupation categories.	
	  }
	\vspace{-2ex}
  \label{fig:occupation_number}
\end{figure}

The PISC dataset consists of 22,670 images. 
The average number of people per image is 3.11. 
For the social relationships,
we consider each individual pair as one sample.
In total, 
we collected 76,568 valid samples.
The distribution for each types of relationships and their agreement rate is shown in \fig~\ref{fig:relation_number}.
The agreement rate is calculated by dividing the number of correct human judgments (judgments that agree with the majority) with the total number of judgments.
For occupations, 
10,034 images contain people that have recognizable occupations.
In total, 
there are 66 identified occupation categories. 
%In this work, 
%we keep the categories that occur more than 200 times for training set and test set, 
%which leaves us with 26 occupations and 23,510 samples.
The number of occupation occurrence and the workers' agreement rate for the 26 most frequent occupation categories are shown in \fig~\ref{fig:occupation_number}.
A lower agreement rate indicates that the occupation is harder to visually discriminate (\eg~`politician' and `office clerk').
Since two source datasets,
\ie~MS-COCO and Visual Genome, 
are highly biased towards `baseball player' and `skier', 
we limit the total number of instances per occupation to 2000 based on agreement rate ranking to ensure there are no bias towards any particular occupation.
% which removed 641 images that containing these two occupations.
%  based on agreement rate ranking, so that the PISC dataset is not unbalanced towards any particular occupation.
\begin{figure*}[!t]
% 	\centering
% 	\begin{minipage}{2.0\columnwidth}
	\centerline{\includegraphics[trim=5 130 5 0,clip,width=1.0\textwidth]{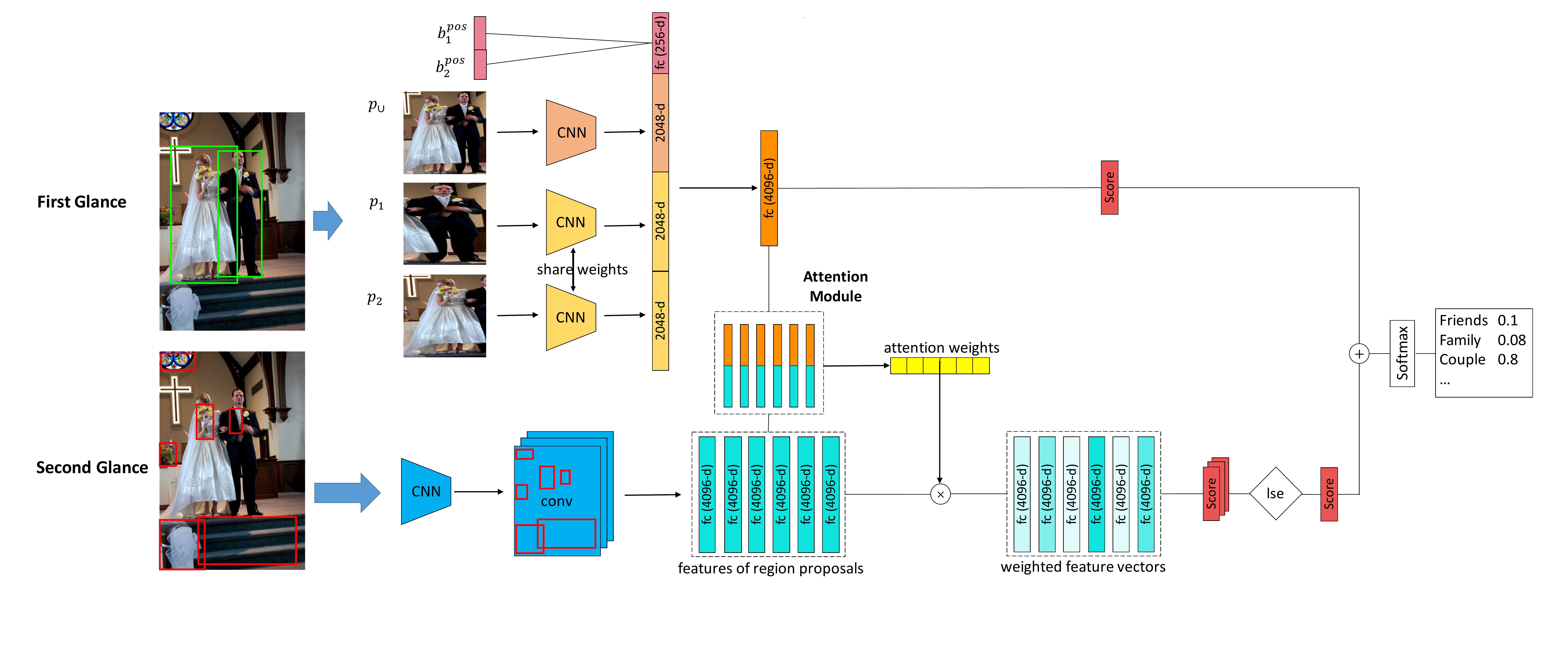}}
% 	\end{minipage}		
  \vspace{-2ex}
  \caption
  	{
  	\small
  	An overview of the proposed dual-glance model. 
  	The first glance looks at the pair of people in question and makes a coarse prediction. The second glance looks at region proposals,
  	allocates attention to each region, and aggregates their outputs to refine the score.
  	The attention is guided by both top-down signal from the first glance, and bottom-up signal form the local context.
  	}
  \label{fig:network}
\end{figure*}
\section{Proposed Dual-Glance Model}
\label{sec:model}

Given an image {\small $\Mat{I}$} and a target pair of people highlighted by bounding boxes {\small$\{b_1,b_2\}$}, 
our goal is to infer their social relationship {\small$r$}.
In this work, 
we propose a dual-glance relationship recognition model,
where the first glance fixates at $b_1$ and $b_2$, and the second glance explores contextual cues from multiple region proposals {\small $\Vec{R}$}.
The final score over possible relationships, $\Vec{S}$, is a weighted sum of the two scores via
\begin{equation}
	\Vec{S} = \Vec{S}_1(\Mat{I},b_1,b_2)+\alpha \Vec{S}_2(\Mat{I},b_1,b_2,\Vec{R}).
\end{equation}

We use softmax to transform the final score into a probability distribution.
Specifically, the probability that a given pair of people having relationship $r$ is given as
\begin{equation}
	p_r = \dfrac{\exp(S_r)}{\sum_{r}\exp(S_r)}.
\end{equation}

An overview of the proposed model is shown in \fig~\ref{fig:network}.

\subsection{First Glance}

The first glance takes in input {\small $\Mat{I}$} and two bounding boxes.
We first crop three patches from {\small $\Mat{I}$}, 
where the first two cover each person, {\small $\Mat{p}_{1}$} and {\small $\Mat{p}_{2}$}, and one for the union region, {\small $\Mat{p}_{\cup}$}, that tightly covers both people.
These patches are resized to {\small $224\times224$} pixels and fed into three CNNs,
The outputs from the last convolutional layer are flattened and concatenated.
{\small $\Mat{p}_{1}$} and {\small $\Mat{p}_{2}$} are processed by CNNs that share the same weights.

We denote the geometry feature of the bounding box {\small $i$} as
{\small $\Vec{b_{\text{i}}^{\text{pos}}} = \{x_{i}^{min},y_{i}^{min},x_{i}^{max},y_{i}^{max},area_{i}\} \in \mathbb{R}^5 $},
where all the parameters are relative values, normalized with zero mean and unit variance.
$\small{\Vec{b_{\text{1}}^{\text{pos}}} }$ and $\small{\Vec{b_{\text{2}}^{\text{pos}}} }$ are concatenated and processed by a fully-connected (fc) layer. 
We concatenate its output with the CNN features for {\small $\Mat{p}_{1}$}, 
{\small $\Mat{p}_{2}$} and {\small $\Mat{p}_{\cup}$} to form a single feature vector,
which is subsequently passed through another two fc layers to produce first glance score, $\Vec{s}_1$.
We use {\small $\Vec{v}_{\text{top}}\in\mathbb{R}^{k}$} to denote the output from the penultimate fc layer.
{\small $\Vec{v}_{\text{top}}$} serves as a top-down signal to guide the attention mechanism in the second glance.
We experimented with different values of {\small $k$},
and set {\small $k$} as 4096.

\subsection{Attentive RCNN for Second Glance}

For the second glance, we adapt Faster RCNN~\cite{Ren_NIPS_2015} to make use of multiple contextual regions.
Faster RCNN processes the input image {\small $\Mat{I}$} with Region Proposal Network (RPN) to generate a set of region proposals {\small $\Vec{P}_{\Mat{I}}$} with high objectness. 
For each target pair with bounding boxes $b_1$ and $b_2$,
we select the set of contextual regions {\small $\Vec{R}(b_1,b_2;\Mat{I})$} from {\small $\Vec{P}_{\Mat{I}}$} as
\begin{equation}
% \begin{split}
	\resizebox{0.90\linewidth}{!}{$
	\Vec{R}(b_1,b_2;\Mat{I}) = \{c \in \Vec{P}_{\Vec{I}}: \max(G(c,b_1),G(c,b_2)) < \tau_u \}
	$}
% \end{split}
\end{equation}

\noindent
where {\small $G(b_1,b_2)$} computes the Intersection-over-Union (IoU) between two regions,
and {\small $\tau_u$} is the upper threshold for IoU.
The threshold encourages the second glance to explore cues different from the first glance.
It's effect is reported in Section~\ref{sec:tau}.

We then process {\small $\Mat{I}$} with a CNN to generate a convolutional feature map \textit{conv}($\Mat{I}$).
For each contextual region {\small $c\in \Vec{R}$}, 
ROI pooling is applied to extract a fixed-length feature vector {\small $\Vec{v}\in\mathbb{R}^{k}$} from \textit{conv}($\Mat{I}$).
Denote {\small $\{\Vec{v}_i|i=1,2,\ldots,N\}$} as the bag of $N$ feature vectors for {\small $\Vec{R}$},
also given the high-level feature vector from the first glance {\small $\Vec{{v}_{\text{top}}}$},         
we first combine them into a hidden vector {\small $\Vec{h}_i\in\mathbb{R}^{k}$} via
\begin{equation}
	\Vec{h}_i = \Vec{v}_i+\Vec{w_{\text{top}}}\otimes \Vec{v_{\text{top}}}, \\	
\end{equation}

\noindent
where {\small $\Vec{{w}_{\text{top}}}\in\mathbb{R}^{k}$}, and {\small $\otimes$} is the element-wise multiplication of two vectors.
Then, we calculate the attention $a_i\in [0,1]$ over the $i$th region proposal as 
\begin{equation}
	\begin{split}
	a_i &= \dfrac{1}{1+\exp(-( \Mat{W}_{h,a} \Vec{h}_i+b_a))},
	\end{split}
\end{equation}

\noindent
where {\small $\Mat{W}_{h,a}\in\mathbb{R}^{1\times k}$} is the weight matrix, and {\small $b_a\in\mathbb{R}$} is the bias term.
The attention over each contextual region is guided by both bottom-up signal from local region {\small $\Vec{v}_i$} and top-down signal from the first glance model {\small$\Vec{v}_{\text{top}}$}.
Hence, 
the weighted feature vector for region $i$ is computed via
\begin{equation}
	\Vec{v}_i^{att} = a_i\Vec{v}_i.
\end{equation}

The obtained $\Vec{v}_i^{att}$ is processed by the last fc layer to generate the output score for the $i$th region proposal  
\begin{equation}
	\Vec{s}_i = \Mat{W}_s\Vec{v}_i^{att}+\Vec{b}_{s}. 
\end{equation}

Functions to aggregate scores,
\ie,~{\small $f(\Vec{s}_i)$}, 
from a bag of instances {\small $i=1,2,\ldots,N$} include {\small $\max(\Vec{s}_i)$, $\text{avg}(\Vec{s}_i)$}, and {\small $\log[1+\sum_{i=1}^{N}{\exp(\Vec{s}_i)}]$} (log-sum-exp, denoted as {\small $\text{lse}(\Vec{s}_i)$}).
In our experiment (Section~\ref{sec:att}),
we evaluated all three variants of {\small $f(\Vec{s}_i)$} and the results show that {\small $\text{lse}(\Vec{s}_i)$} provides the best performance.
Hence, 
the score of the second glance model is
\begin{equation}
	\Vec{S}_2 = \log[1+\sum\nolimits_{i=1}^{N}{\exp(\Vec{s}_i)}]. 
\end{equation}

\section{Experiment}
\label{sec:experiment}

\subsection{Dataset and Training Details}

In this work, 
we conducted experiments on the proposed PISC dataset (Section~\ref{sec:dataset}) and evaluated our proposed method with two recognition tasks.
The first task,
denoted as 3-relationship recognition,
focuses on three coarse-level relationship categories,
namely {\it No Relation}, {\it Intimate Relation}, and {\it Non-Intimate Relation}.
We randomly select 4,000 images (14,852 samples) as test set,
and use the remaining images as training set.
The second task,
denoted as 6-relationship recognition,
focuses on finer relationships listed in \fig~\ref{fig:social_relationship_tree}.
Since the data label is unbalanced (fewer images with {\it Couple} or {\it Commercial relationship}),
we split 1,500 images (3,517 samples) into test set and ensure it contains around 600 samples for each of the six relationships.
%The remaining images are used as training set.

The relationship imbalance reflects their frequency of occurrence, which is also observed in~\cite{Lu_ECCV_2016}.
%The data in the training set for 6-relationship recognition has the label bias problem.
To address this,
we adopt oversampling and undersampling strategies.
Specifically,
we oversample the minority labeled samples by reversing the pair of people 
(\ie,~if $p_1$ and $p_2$ are a couple, then $p_2$ and $p_1$ are also a couple),
and by horizontally flipping the image.
We undersample the majority labeled samples using stratified sampling scheme to ensure the samples in each batch is balanced.

%We undersample the majority samples by dropping a sample during training with a probability
%$p_\text{d}^\text{r} = N_\text{comm}/N_\text{r}$, where $N_\text{comm}$ is the total number of %training samples for commercial relationship (fewest samples),
%and $N_\text{r}$ is the total number of training samples for majority relationship $r$. 
%\NOTE{rewrite the last sentence as step sampling}

In this work, 
we train our model with Stochastic Gradient Descent using backpropagation.
First, 
we train the first-glance model until the loss converges,
then we freeze the first-glance model, and train the second-glance model.
For the first glance model, we fine-tune the ResNet-101 model pre-trained on ImageNet classification task~\cite{He_2016_CVPR}.
For the second glance model, we fine-tune the VGG-16 model pre-trained on ImageNet detection task~\cite{Ren_NIPS_2015}. 
We set the learning rate as 0.001, while the fine-tuning model has a lower learning rate of 0.0001.
We use a batch size of 32 and a momentum of 0.9 during training.

During the test stage, 
we found that the performance would slightly improve if we feed the model twice with $\{b_1, b_2\}$ and $\{b_2, b_1\}$, and take their average as the final score.
However, 
the performance gain (0.5\%-1\%) doubles the time budget, 
and we do not recommend it in practice.

\begin{table*}[!t]
	\centering
	\caption
		{
		\small	
	  Recall-per-class and mean average precision (mAP) of baselines and our proposed dual-glance model on the PISC dataset. 
		}
	\label{tbl:fullresult}
	\vspace{-2ex}
	\small	
	\begin{tabular}{l|c|c|c|c||c|c|c|c|c|c|c} 
	  \toprule
		  & \multicolumn{4}{c||}{3-relationship} &	\multicolumn{7}{c}{6-relationship}		  \\
	  \cmidrule{2-12}
			& \begin{sideways} \parbox{1.8cm}{\raggedright Intimate} 				\end{sideways} 
		  & \begin{sideways} \parbox{1.8cm}{\raggedright Non-Intimate} 		\end{sideways} 
		  & \begin{sideways} \parbox{1.8cm}{\raggedright No Relation} 		\end{sideways} 
		  & \begin{sideways} \parbox{1.8cm}{\raggedright mAP} 	\end{sideways} 
		  & \begin{sideways} \parbox{1.8cm}{\raggedright Friends} 				\end{sideways} 
		  & \begin{sideways} \parbox{1.8cm}{\raggedright Family} 	\end{sideways} 
		  & \begin{sideways} \parbox{1.8cm}{\raggedright Couple} 					\end{sideways} 
		  & \begin{sideways} \parbox{1.8cm}{\raggedright Professional} 		\end{sideways} 
		  & \begin{sideways} \parbox{1.8cm}{\raggedright Commercial } 			\end{sideways} 
		  & \begin{sideways} \parbox{1.8cm}{\raggedright No Relation} 		\end{sideways} 
			& \begin{sideways} \parbox{1.8cm}{\raggedright mAP} 	\end{sideways} \\
	  \midrule
	  Union-CNN~\cite{Lu_ECCV_2016} & 72.1 & 81.8 & 19.2 & 58.4 &29.9 & 58.5 & 70.7 &55.4 &43.0 &19.6 &43.5 \\
	  BBox & 42.4 & 33.0  & 41.9 & 34.9 & 20.7 & 36.4 & 42.7 & 31.8& 23.2 & 32.7 & 28.8 \\	  	  	  	  
	  Pair-CNN & 70.3 & 80.5  & 38.8 & 65.1 & 30.2 & 59.1  & 69.4  &57.5 & 41.9 &34.2  & 48.2 \\	  	  	  	  
	  Pair-CNN+BBox & 71.8 & 80.3 & 50.6& 69.6 &30.7 & 60.2 & 72.5   &58.1  &43.7 &50.7 & 54.3\\		  	    	 
		Pair-CNN+BBox+Union & 71.1& 81.2 & {\bf 57.9} & \bf{72.2} & {\bf 32.5} & {\bf 62.1} & {\bf 73.9} & {\bf 61.4} & {\bf 46.0} & {\bf 52.1} & \bf{56.9}\\
		Pair-CNN+BBox+Global & 70.5 & 80.9 & 53.7& 70.5 & 32.2& 61.7 & 72.6 & 60.8&44.3 &51.0 & 54.6\\		  
		Pair-CNN+BBox+Scene  &71.0 & 80.6 & 46.7 & 68.0 &30.2 &59.4 &71.7 &57.6 &43.0 &49.9 & 51.7\\
		RCNN  & {\bf 72.9} & {\bf 83.3} & 14.8& 63.5 & 29.7& 61.9 & 71.2 & 60.1&45.9 &20.7 & 48.4\\	\midrule
		Dual-Glance & {\color{blue} {\bf 73.1}}& {\color{blue} {\bf 84.2}}& {\color{blue} {\bf59.6}}& {\color{blue} {\bf 79.7}} &{\color{blue} {\bf35.4}} & {\color{blue} {\bf68.1}}& {\color{blue} {\bf76.3}} & {\color{blue} {\bf70.3}} &{\color{blue} {\bf57.6}} & {\color{blue} {\bf60.9}}& {\color{blue} {\bf 63.2}}\\	  
	  \bottomrule
	\end{tabular}
\end{table*}

\subsection{Single-Glance vs. Dual-Glance}

As there exists limited literature on this problem,
we evaluate multiple variants of our model as baseline and compare them to the proposed dual-glance mode to show its efficacy.
Formally, 
the compared methods are as followed:
% In addition, 
% we explore the effect of box location, local context and scene context,
% as well as demonstrate the efficacy of the proposed dual-glance model quantitatively and qualitatively.

\begin{enumerate}
  \item 
% \noindent
\textbf{Union-CNN}: 
Following the predicate prediction model in~\cite{Lu_ECCV_2016},
a single CNN model is used to classify the union region of the individual pair of interest.
% 
% Following the predicate prediction model in~\cite{Lu_ECCV_2016},
% we select the union region of the pair of individuals participating in a relationship and use a single CNN model for classification.
% 
% The model consists of single CNN model. 
% Similar to the predicate prediction model in~\cite{Lu_ECCV_2016}, the input is the union region of the pair of individuals participating in a relationship.
\item
% \noindent
\textbf{BBox}:
We only use the geometry feature of the two bounding boxes to infer the relationship.
\item
% \noindent
\textbf{Pair-CNN}: 
The model consists of two CNNs with shared weights. The input is the cropped image patches for the two individuals.
\item
% \noindent
\textbf{Pair-CNN+BBox}:
We extend Pair-CNN by using the geometry feature of the two bounding boxes.
\item
% \noindent
\textbf{Pair-CNN+BBox+Union}: 
The first glance model as illustrated in \fig~\ref{fig:network}, which combines Pair-CNN+BBox and Union-CNN.
\item
% \noindent
\textbf{Pair-CNN+BBox+Global}:
Instead of the union region, we use the entire image as input to Union-CNN.  
\item
% \noindent
\textbf{Pair-CNN+BBox+Scene}:
The Union-CNN is replaced with a Scene-CNN pre-trained on Places~\cite{Zhou_CoRR_2016}. It extracts scene information using the entire image.
\item
% \noindent
\textbf{RCNN}:
We train a RCNN using the region proposals {\small $\Vec{P}_{\Mat{I}}$}, and adopt average pooling to combine the features.
\item
% \noindent
\textbf{Dual-Glance}:
Our proposed model (Section~\ref{sec:model}). 
\end{enumerate}

%\noindent
%\textbf{Variants of Dual-Glance}:
%See Section~\ref{sec:att}.

\tab~\ref{tbl:fullresult} shows the results on the test set for both the 3-relationship recognition task and the 6-relationship recognition task.
\textbf{Union-CNN} and \textbf{RCNN}, are incapable to recognize \textit{No Relation}.
This is because these model don't know the pair of people in question,
and would recognize other salient relationships.
\textbf{Pair-CNN+BBox} outperforms \textbf{Pair-CNN}, which suggests that peoples' geometric position in an image contains information useful to infer their relationship, especially for \textit{No Relation}. 
This is supported by the law of \textit{proxemics} defined in the book "The Silent Language"~\cite{Hall_1959_silent}.
However, the position of bounding boxes alone cannot be used to predict relationship, as shown by the results of \textbf{BBox}.

Adding \textbf{Union-CNN} to \textbf{Pair-CNN+BBox} improves performance. 
However, the performance gain is slight if we use the global context (entire image) rather than local context (union region).
Furthermore, 
the performance even degrades when the global context incorporates scene information,
suggesting that social relationships are independent of scene types.
\textbf{RCNN} demonstrates the effectiveness of using contextual regions,
particularly for {\it Intimate Relation} and {\it Non-Intimate Relation}.

\begin{figure}[!t]
  \centering
%   \begin{minipage}{1.0\columnwidth}
%   	\begin{minipage}{1.0\columnwidth}
%   		\begin{minipage}{0.495\columnwidth} \centerline{\includegraphics[width=\linewidth]{example_dual_1}} \end{minipage}
%   		\begin{minipage}{0.495\columnwidth} \centerline{\includegraphics[width=\linewidth]{example_dual_3}} \end{minipage}  	
%   	\end{minipage}
%   	\begin{minipage}{1.0\columnwidth}
%   		\begin{minipage}{0.495\columnwidth} \centerline{\includegraphics[width=\linewidth]{example_dual_2}} \end{minipage}
%   		\begin{minipage}{0.495\columnwidth} \centerline{\includegraphics[width=\linewidth]{example_dual_4}}	\end{minipage}  	
%   	\end{minipage}
%   \end{minipage}
	\begin{minipage}{1.0\columnwidth}
		\centerline{\includegraphics[width=1.0\linewidth]{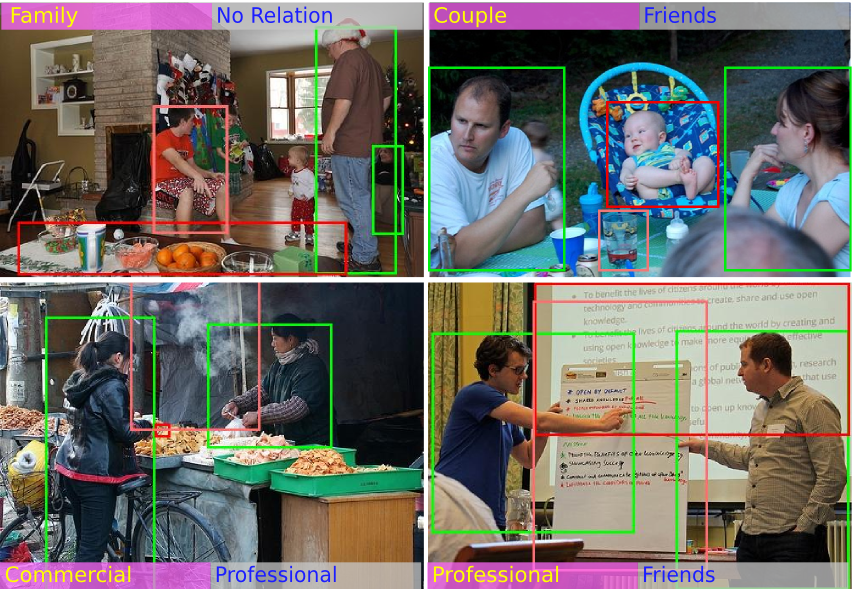}}
	\end{minipage}  
  \vspace{-2ex}
  \caption
    {
    \small
    Examples where dual-glance model correctly predict the relationship (yellow label) while the first-glance model fails (blue label). 
     \textcolor{green}{GREEN} boxes highlight the pair of people in question,
    and the top two contextual regions with highest attention are highlighted in \textcolor{red}{RED}.
    }
   	\vspace{-2ex} 
  \label{fig:dual_correct}
  	\vspace{-1ex} 
\end{figure}

The proposed \textbf{Dual-Glance} significantly outperforms all baseline models.
%Comparing the dual-glance model with the first-glance model (Pair-CNN+BBox+Union),
%the performance gain is highest for \textit{Commercial} (+9.6) and \textit{No Relation} (+9.8), while lower for \textit{Couple} (+1.2).
\fig~\ref{fig:dual_correct} shows some intuitive illustrations where proposed model correctly classifies relationships misclassified by the first-glance (\textbf{Pair-CNN+BBox+Union}) model.

\begin{figure}[!t]
  \centering
%   \begin{minipage}{1.0\columnwidth}
%   	\begin{minipage}{0.465\columnwidth} \centerline{\includegraphics[trim=60 10 90 0,clip,width=1.0\linewidth]{FG_6class_confusion_mat}} 	\end{minipage}
%   	\begin{minipage}{0.525\columnwidth} \centerline{\includegraphics[trim=60 10 40 0,clip,width=1.0\linewidth]{DG_6class_confusion_mat}}	\end{minipage}
%   \end{minipage}
	\begin{minipage}{1.0\columnwidth}
		\centerline{\includegraphics[width=1.0\linewidth]{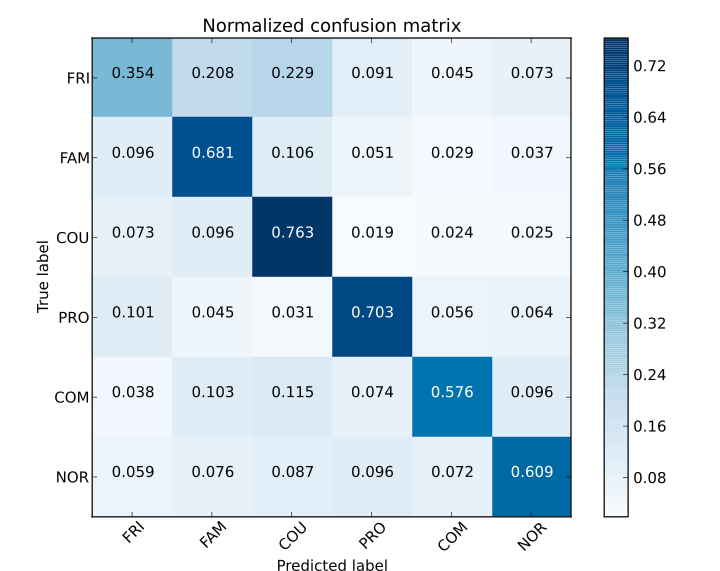}}
	\end{minipage}  
  \vspace{-2ex}
  \caption
    {
    \small
    Confusion matrix of 6-relationship recognition task with the proposed dual-glance model.       
    }
    	\vspace{-2ex} 
  \label{fig:confusion_dual_6class}
\end{figure}

Across all models, \textit{Friends} and \textit{Commercial} are more difficult to recognize.
This is consistent with the agreement rate in Figure~\ref{fig:relation_number},
which indicates that \textit{Friends} and \textit{Commercial} are less visually distinguishable.
\fig~\ref{fig:confusion_dual_6class} shows the confusion matrix of 6-relationship recognition task.
The three intimate relationships (\textit{Friends}, \textit{Family}, \textit{Couple}) are more often to be confused with each other than with non-intimate relationships, 
suggesting they share similar visual features.
However, the non-intimate relationships (\textit{Professional}, \textit{Commercial}) do not tend to be easily confused with each other.

\subsection{Analysis on Attention Mechanism}
\label{sec:att}

Here,
we remove the attention module and compare it with our proposed dual-glance model.
For the second glance, we experiment with three widely used aggregation functions $f(\cdot)$, 
which are $\text{avg}(\cdot)$, $\text{lse}(\cdot)$ and $\max(\cdot)$.
The results are shown in \tab~\ref{tbl:MIL}.
Adding attention mechanism improves performance for all three aggregation functions.
For Dual-glance without attention, $\max(\cdot)$ performs best, which conforms to the results in~\cite{Gkioxari_ICCV_2015,Wu_CVPR_2015}.
While for Dual-glance with attention, $\text{lse}(\cdot)$ performs best.

The reason is that $\max(\cdot)$ uses a single instance to infer the label for an entire bag.
It works well in the presence of a `strong' instance,
but sometimes there is no strong instance, but several `weak' instances.
On the other hand,
$\text{lse}(\cdot)$ and $\text{avg}(\cdot)$ consider all instances in a bag,
but could be distracted by irrelevant instances.
However, with properly guided attention, 
$\text{lse}(\cdot)$ and $\text{avg}(\cdot)$ can better exploit the collaborative power of relevant instances for more accurate inference.
  
\subsection{Variations of Contextual Regions}
\label{sec:tau}
\begin{table}[!t]
	\centering
	\caption
		{
		\small	
		mAP (\%) of the proposed dual-glance model with and without attention mechanism using various aggregattion functions.
		}
	\label{tbl:MIL}
	\vspace{-2ex}
	\small
	\begin{tabularx}{1\columnwidth}{l|c|c|c||c|c|c} 
		\toprule
		 &	\multicolumn{3}{c||}{Without Attention} &	\multicolumn{3}{c}{With Attention}
		\\
		 & \!{\scriptsize $\text{avg}(\cdot)$}\!  & \!{\scriptsize $\text{lse}(\cdot)$}\! & \!\!{\scriptsize $\max(\cdot)$}\!\! 
		 & \!{\scriptsize $\text{avg}(\cdot)$}\!  & \!{\scriptsize $\text{lse}(\cdot)$}\! & \!\!{\scriptsize $\max(\cdot)$}\!\! \\
	  \midrule
		\!\! 3-relationship  &71.7&73.0 &74.8 &76.9 &\bf{79.7} & 77.6   \\
		\!\! 6-relationship  &55.9&57.5 &58.2 &61.8 &\bf{63.2} & 62.1   \\	
	  \bottomrule
	\end{tabularx}
\end{table}	
\begin{figure}[!t]
  \centering
	\begin{minipage}{1.0\columnwidth}
		\begin{minipage}{0.49\columnwidth} \centerline{\includegraphics[width=1.0\linewidth]{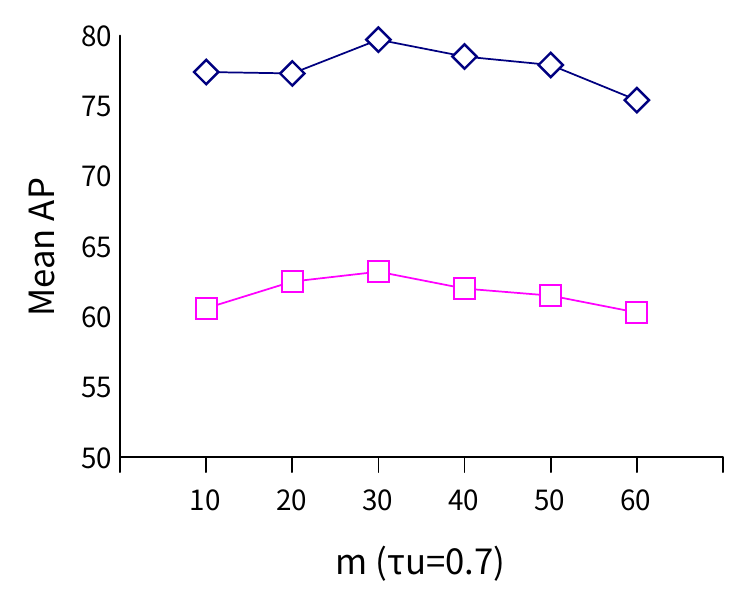}} \end{minipage}
		\begin{minipage}{0.49\columnwidth} \centerline{\includegraphics[width=1.0\linewidth]{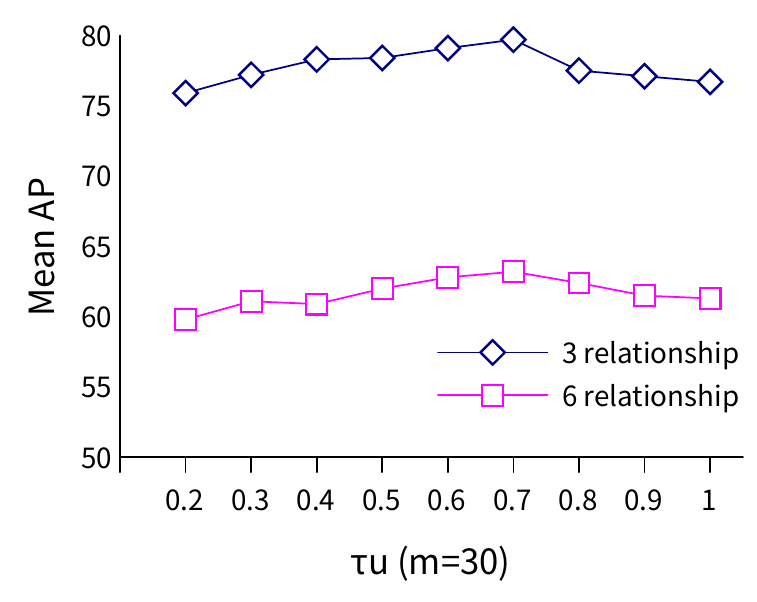}} \end{minipage}
	\end{minipage}
  \vspace{-2ex}
  \caption
    {
    \small
		Evaluation of dual-glance model over variations in maximum number of region proposals (Left) and upper threshold of overlap between region proposals and the pair of people (Right).
    }
    \vspace{-2ex} 
  \label{fig:regions}
  
\end{figure}

Since RPN can generate hundreds of region proposals per image,
we suppress those proposals with non-maximum suppression (NMP).
We vary $m$, the maximum number of region proposals used, as well as $\tau_u$, the upper threshold of overlap between a region proposal and the target people.
We experimented with different combinations of $m$ and $\tau_u$ with the dual-glance model.
As shown in \fig~\ref{fig:regions}, $m=30$ and $\tau_u=0.7$ produce the best performance.

\begin{figure}[!t]
	\centering
	\begin{minipage}{1.0\columnwidth}
		\centerline{\includegraphics[width=\columnwidth]{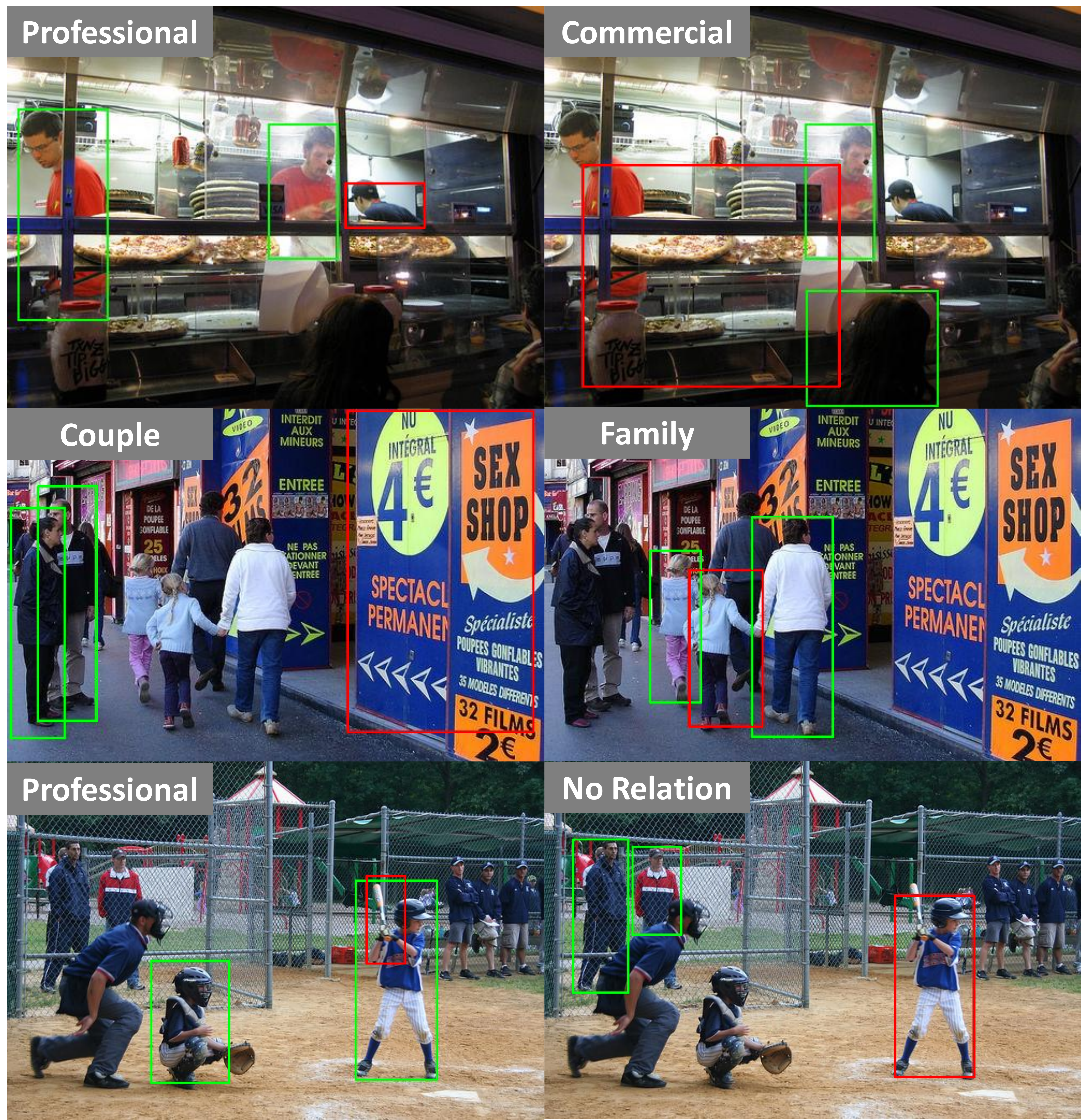}}
	\end{minipage}
	\vspace{-2ex}  
  \caption
    {
    \small
    Illustration of the proposed attentive RCNN.
    \textcolor{green}{GREEN} boxes highlight the pair of people in question, and \textcolor{red}{RED} box highlights the context region with the highest attention.
    For each target pair, the attention mechanism fixates on different region.  
    }
  \label{fig:visualize_attention}
  \vspace{-2ex} 
\end{figure}

\subsection{Visualization of Examples}
\label{sec:visualization}

\begin{figure*}[!t]
  \centering
   %\begin{minipage}{1.0\textwidth}
   	\begin{minipage}{1.0\textwidth}
   		\centerline{\includegraphics[width=1.0\textwidth]{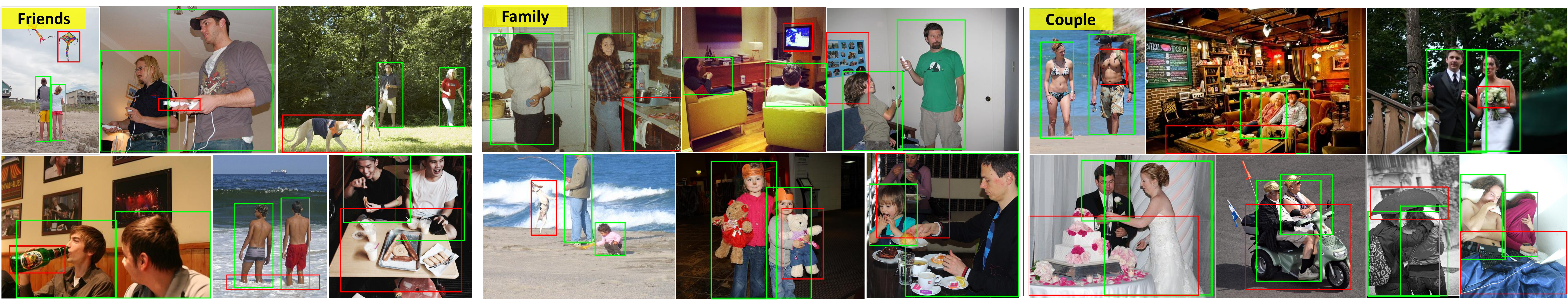}}
   		\vspace{-1ex}
   		\centerline{\small Intimate Relation}
%    		\vspace{1ex}
   	\end{minipage}	
 		\begin{minipage}{1.0\textwidth}
 			\begin{minipage}{.68\textwidth}
 	  		\centerline{\includegraphics[width=1.0\linewidth]{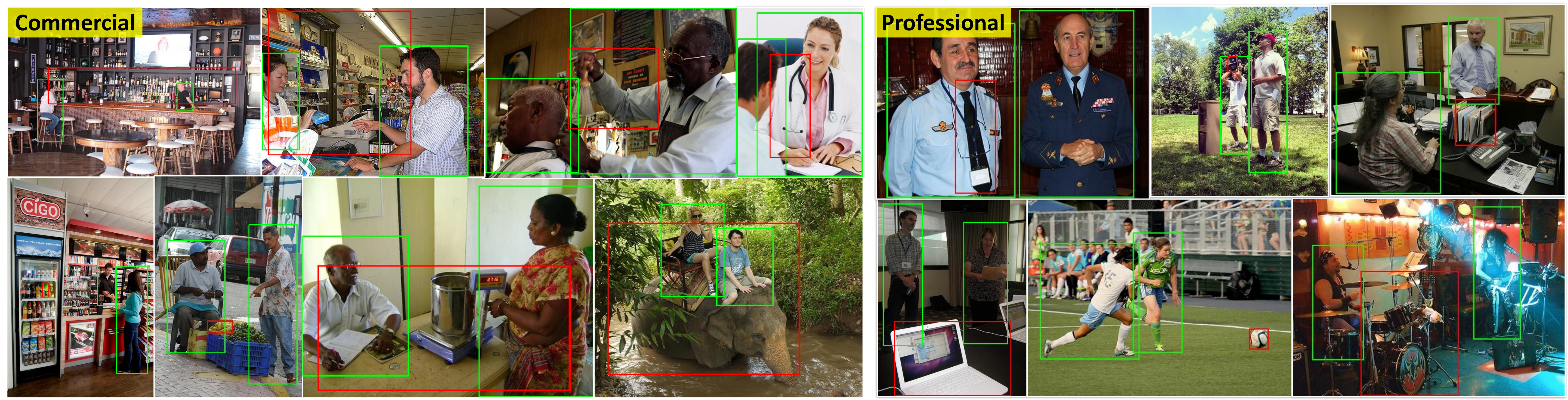}}
 	  		\vspace{-1ex}
 	  		\centerline{\small Non-Intimate Relation}
 	  	\end{minipage}
 	  	%\hfill
 	  	\begin{minipage}{.32\textwidth}
   			\centerline{\includegraphics[width=1.0\linewidth]{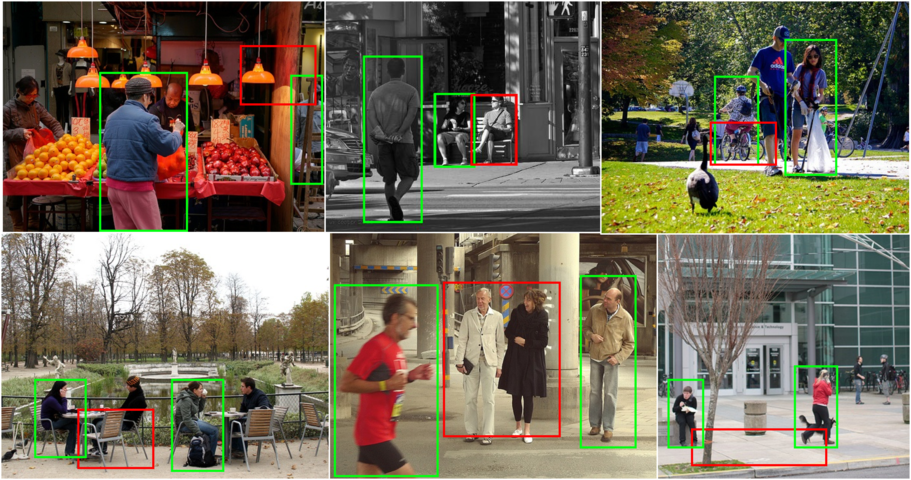}}
   			\vspace{-1ex}
   			\centerline{\small No Relation}
   		\end{minipage}
 		\end{minipage}
 	%\end{minipage}  	
% 
%	\begin{subfigure}[b]{0.7\textwidth} 		\includegraphics[width=1\textwidth]{example-friends-small}
%\end{subfigure}  
%	\begin{subfigure}[b]{0.7\textwidth}
%		\includegraphics[width=1\textwidth]{example-family-small}
%	\end{subfigure}  
%	\begin{subfigure}[b]{0.7\textwidth}
%		\includegraphics[width=1\textwidth]{example-couple-small}
%	\end{subfigure}  
%	\begin{subfigure}[b]{0.7\textwidth}
%		\includegraphics[width=1\textwidth]{example-commercial-small}
%	\end{subfigure}  
%	\begin{subfigure}[b]{0.7\textwidth}
%		\includegraphics[width=1\textwidth]{example-professional-small}
%	\end{subfigure}  
%	\begin{subfigure}[b]{0.7\textwidth}
%		\includegraphics[width=1\textwidth]{example-no-small}
%	\end{subfigure}  
	\vspace{-1ex}
  \caption
    {
    \small
    Example of correct predictions on PISC dataset. Green boxes highlight the targets, and red box highlights the contextual region with highest attention.
    }
  \label{fig:visualize_correct}
%   \vspace{-2ex}
\end{figure*}

\begin{figure}[!t]
	\centering
	\begin{minipage}{1.0\columnwidth}
		\centerline{\includegraphics[width=1.0\linewidth]{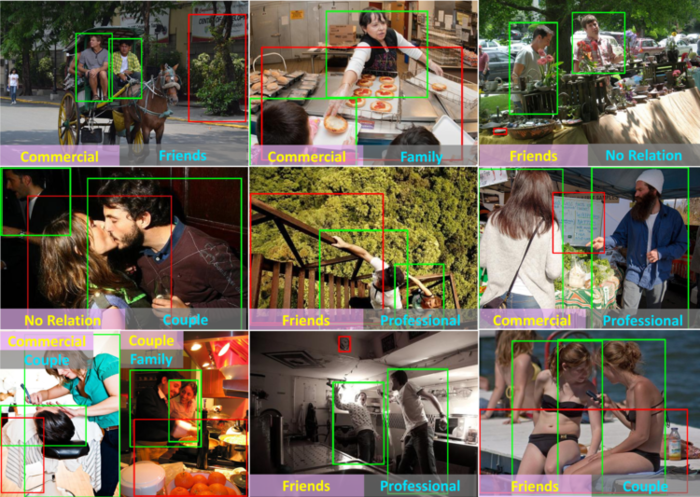}}
	\end{minipage}  
  \vspace{-1ex}
  \caption
    {
    \small
    Examples of incorrection predictions on PISC dataset. 
    Yellow labels are the ground truth, and blue labels are the model's predictions.
    }
  \label{fig:visualize_wrong}
   \vspace{-3ex}
\end{figure}
The attention mechanism enables different pairs of people in question to exploit different contextual cues.
Some examples are shown in \fig~\ref{fig:visualize_attention}.
%The target pair is highlighted in green, and the context region with the highest attention is highlighted in red.
In the second row, the little girl in red box is useful to infer that the other girl on her left and the woman on her right are family members,
but her existence indicates little of the couple in black.

\fig~\ref{fig:visualize_correct} shows examples of correct recognition for each relationship category in the test set. 
We can observe that the proposed model learns to recognize social relationship from a wide range of visual cues including clothing, environment, surrounding people/animals, contexual objects, etc.
For intimate relationships, the contextual cues varies from \textit{beer} (friends),
\textit{gamepad} (friends),
\textit{TV} (family), to \textit{cake} (couple) and \textit{flowers} (couple).
In terms of non-intimate relationships, the contextual cues are related to the occupations of the individuals.
For instance, \textit{goods shelf} and \textit{scale}  indicate commercial relationship, 
while \textit{uniform} and \textit{documents} imply professional relationship.
\fig~\ref{fig:visualize_wrong} shows the misclassified cases. 
The proposed model fails to recognize the gender (misclassifies \textit{friends} as \textit{couple} in the image at row 3 column 3),
or picks up the wrong cue (the white board instead of the vegetable in the image at row 2 column 3).

\section{Conclusion}
\label{sec:conclusion}

In this study, we aim to address pairwise social relationship recognition, 
a key challenge to bridge the social gap towards higher-level social scene understanding.
To this end, we propose a dual-glance model, which exploits useful information from the individual pair of interest as well as multiple contextual regions.
We incorporate attention mechanism to assess the relevance of each region instance with respect to the target pair.
We evaluate the proposed model on PISC dataset, 
a large-scale image dataset we collected to facilitate research in social scene understanding.
We demonstrate both quantitatively and qualitatively the efficacy of the proposed model.
 We also experiment with a few variants of the proposed system to explore information useful for social relationship inference.
 %The models are available at \url{http://anonmous_URL_2.com}.

Our work is the first step towards general social scene understanding in a data-driven fashion.
The PISC dataset provides further potential in this line of research,
including but not limited to group social relation analysis, occupation recognition, and joint inference of social role and social relationships.
We intend to address some of those challenges in future work.
% In addition, we will study dynamic social scene understanding in videos.
 
\section*{Acknowledgment}
\label{sec:acknowledgement}

This research is supported by the National Research Foundation, 
Prime Minister's Office, 
Singapore under its International Research Centre in Singapore Funding Initiative. 
{
\small
\balance
\bibliographystyle{ieee}
\bibliography{bib}
}

\end{document}